\documentclass[11pt,a4paper]{article}
\usepackage[hyperref]{acl2020}
\usepackage{times}
\usepackage{latexsym}

\usepackage{hyperref}
\usepackage{subfig}
\usepackage{mwe}
\usepackage{graphicx}
\usepackage{multirow}
\usepackage{array}
\usepackage{booktabs}
\usepackage{float} 
\usepackage{amssymb}

\usepackage{microtype}

\aclfinalcopy 

\title{Leveraging Monolingual Data with Self-Supervision\\for Multilingual Neural Machine Translation}

\author{Aditya Siddhant, Ankur Bapna, Yuan Cao, Orhan Firat, Mia Chen, \\
\textbf{Sneha Kudungunta, Naveen Arivazhagan, Yonghui Wu} \\
Google Research \\
\{adisid, ankurbpn, yuancao, orhanf, miachen, snehakudugunta, navari,  yonghui\}@google.com}

\date{}

\begin{document}
\maketitle
\begin{abstract}
Over the last few years two promising research directions in low-resource neural machine translation (NMT) have emerged. The first focuses on utilizing high-resource languages to improve the quality of low-resource languages via multilingual NMT. The second direction employs monolingual data with self-supervision to pre-train translation models, followed by fine-tuning on small amounts of supervised data. In this work, we join these two lines of research and demonstrate the efficacy of monolingual data with self-supervision in multilingual NMT. We offer three major results: (i) Using monolingual data significantly boosts the translation quality of low-resource languages in multilingual models. (ii) Self-supervision improves zero-shot translation quality in multilingual models. (iii) Leveraging monolingual data with self-supervision provides a viable path towards adding new languages to multilingual models, getting up to 33 BLEU on WMT ro-en translation without any parallel data or back-translation.
\end{abstract}

\section{Introduction}
Recent work has demonstrated the efficacy of multilingual neural machine translation (multilingual NMT) on improving the translation quality of low-resource languages \cite{firat2016multi,aharoni2019massively} as well as zero-shot translation \cite{ha2016universal,johnson2017google,arivazhagan2019massively}. The success of multilingual NMT on low-resource languages relies heavily on transfer learning from high-resource languages for which copious amounts of parallel data is easily accessible. However, existing multilingual NMT approaches often do not effectively utilize the abundance of monolingual data, especially in low-resource languages. On the other end of the spectrum, self-supervised learning methods, consuming only monolingual data, have achieved great success on transfer learning \cite{devlin2019bert} and unsupervised NMT \cite{lample2018unsupervised,artetxe2018unsupervised} without fully benefiting from the rich learning signals offered by the bilingual data of multiple languages. 

In this work, we propose to combine the beneficial effects of multilingual NMT with the self-supervision from monolingual data. Compared with multilingual models trained without any monolingual data, our approach shows consistent improvements in the translation quality of all languages, with greater than 10 BLEU points improvements on certain low-resource languages. We further demonstrate improvements in zero-shot translation, where our method has almost on-par quality with pivoting-based approaches, without using any alignment or adversarial losses. The most interesting aspect of this work, however, is that we introduce a path towards effectively adding new unseen languages to a multilingual NMT model, showing strong translation quality on several language pairs by leveraging only monolingual data with self-supervised learning, without the need for any parallel data for the new languages.

\begin{figure}[t]
  \centering
\includegraphics[width=1\linewidth]{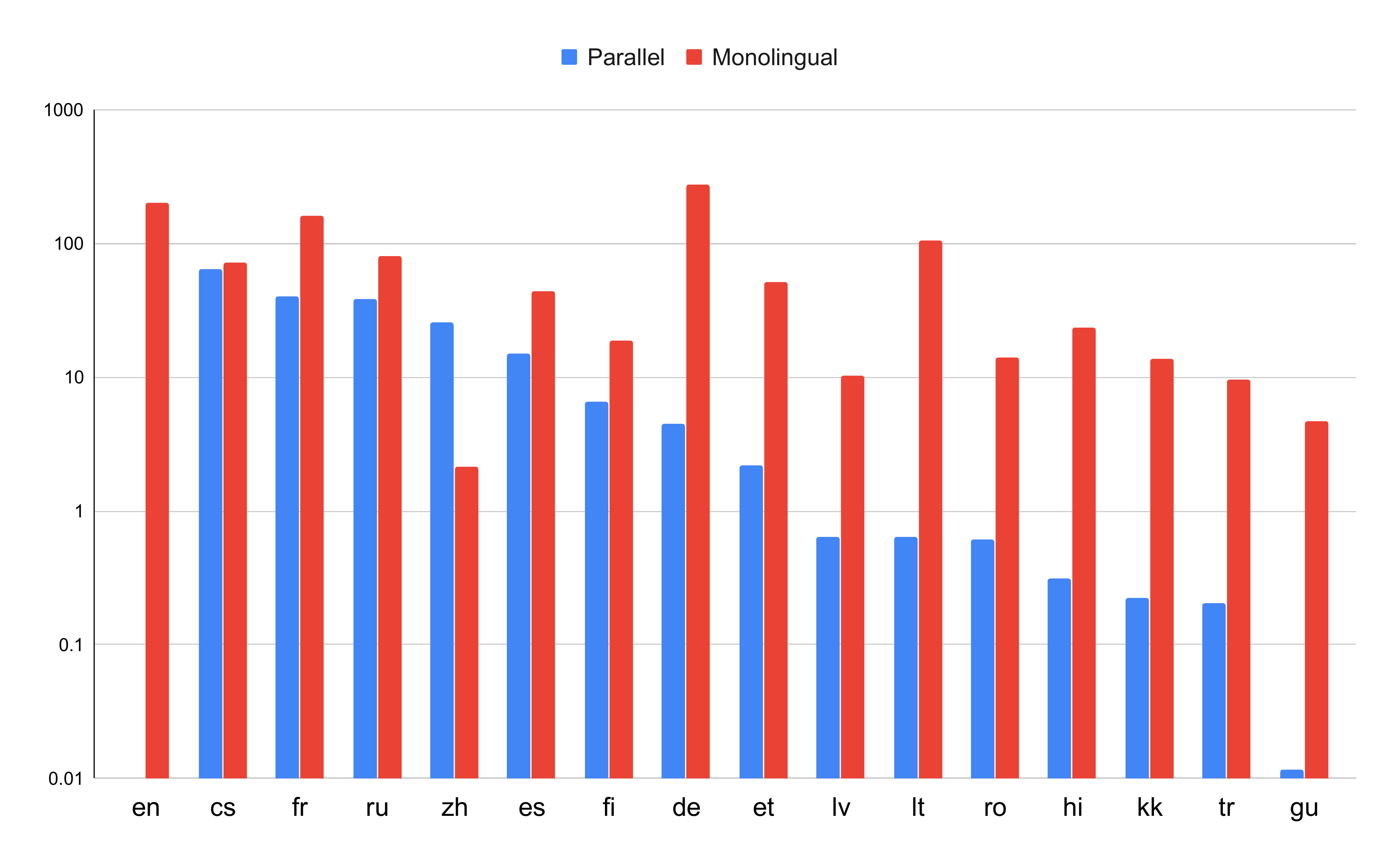}
  \caption{Number of parallel and monolingual training samples in millions for each language in WMT training corpora.}
  \label{fig:datasets}
\end{figure}

\begin{table*}[t]
\centering
\resizebox{\linewidth}{!}{
\begin{tabular}{c|ccccccccccccccc}
\toprule
xx                  & cs   & fr   & ru   & zh   & es   & fi   & de   & et   & lv   & lt   & ro   & hi  & kk  & tr   & gu  \\ \midrule
Any-to-English (xx$\rightarrow$en) & 31.3 & 37.2 & 36.0 & 21.7 & 32.7 & 27.3 & 31.7 & 23.1 & 15.0 & 21.3 & 30.1 & 8.5 & 11.5 & 15.9 & 1.0 \\
English-to-Any (en$\rightarrow$xx) & 23.8 & 41.3 & 26.4 & 31.3 & 31.1 & 18.1 & 29.9 & 18.2 & 14.2 & 11.5 & 23.4 & 4.5 & 1.9  & 13.6 & 0.6 \\ \bottomrule
\end{tabular}} 
\caption{Bilingual baselines. xx refers to language in the column header.} \label{tab:bilingual}
\end{table*}

\section{Method}
We propose a co-training mechanism that combines supervised multilingual NMT with monolingual data and self-supervised learning. While several pre-training based approaches have been studied in the context of NMT \cite{dai2015semi,conneau2019cross,song2019mass}, we proceed with Masked Sequence-to-Sequence (MASS) \cite{song2019mass} given its success on unsupervised and low-resource NMT, and adapt it to the multilingual setting.

\subsection{Adapting MASS for multilingual models}
MASS adapts the masked de-noising objective \citep{devlin2019bert,raffel2019exploring} for sequence-to-sequence models, by masking the input to the encoder and training the decoder to generate the masked portion of the input. To utilize this objective function for unsupervised NMT, \citet{song2019mass} enhance their model with additional improvements, including language embeddings, target language-specific attention context projections, shared target embeddings and softmax parameters and high variance uniform initialization for target attention projection matrices\footnote{Verified from open-source Github implementation.}.

We use the same set of hyper-parameters for self-supervised training as described in \cite{song2019mass}. However, while the success of MASS relies on the architectural \textit{modifications} described above, 
we find that our multilingual NMT experiments are stable even in the absence of these techniques, thanks to the smoothing effect of multilingual joint training. We also forego the separate source and target language embeddings in favour of pre-pending the source sentences with a $<2xx>$ token \cite{johnson2017google}.

We train our models simultaneously on supervised parallel data using the translation objective and on monolingual data using the MASS objective. To denote the target language in multilingual NMT models we prepend the source sentence with the $<2xx>$ token denoting the target language.

\section{Experimental Setup}

\subsection{Datasets}
We use the parallel and monolingual training data provided with the WMT corpus, for 15 languages to and from English. The amount of parallel data available ranges from more than 60 million sentence pairs as in En-Cs to roughly 10k sentence pairs as in En-Gu. We also collect additional monolingual data from WMT news-crawl, news-commentary, common-crawl, europarl-v9, news-discussions and wikidump datasets in all 16 languages including English.\footnote{Followed the versions recommended by WMT'19 shared task, as in http://statmt.org/wmt19/translation-task.html} The amount of monolingual data varies from 2 million sentences in Zh to 270 million in De. The distribution of our parallel and monolingual data is depicted in Figure~\ref{fig:datasets}.

\subsection{Data Sampling}
Given the data imbalance across languages in our datasets, we use a temperature-based data balancing strategy to over-sample low-resource languages in our multilingual models \citep{arivazhagan2019massively}. We use a temperature of $T=5$ to balance our parallel training data. When applicable, we sample monolingual data uniformly across languages since this distribution is not as skewed. For experiments that use both monolingual and parallel data, we mix the two sources at an equal ratio (50\% monolingual data with self-supervision and 50\% parallel data). 

\begin{figure*}[h]
  \centering
  \subfloat[Any-to-English (xx $\rightarrow$ en) ]{\includegraphics[width=0.49\textwidth]{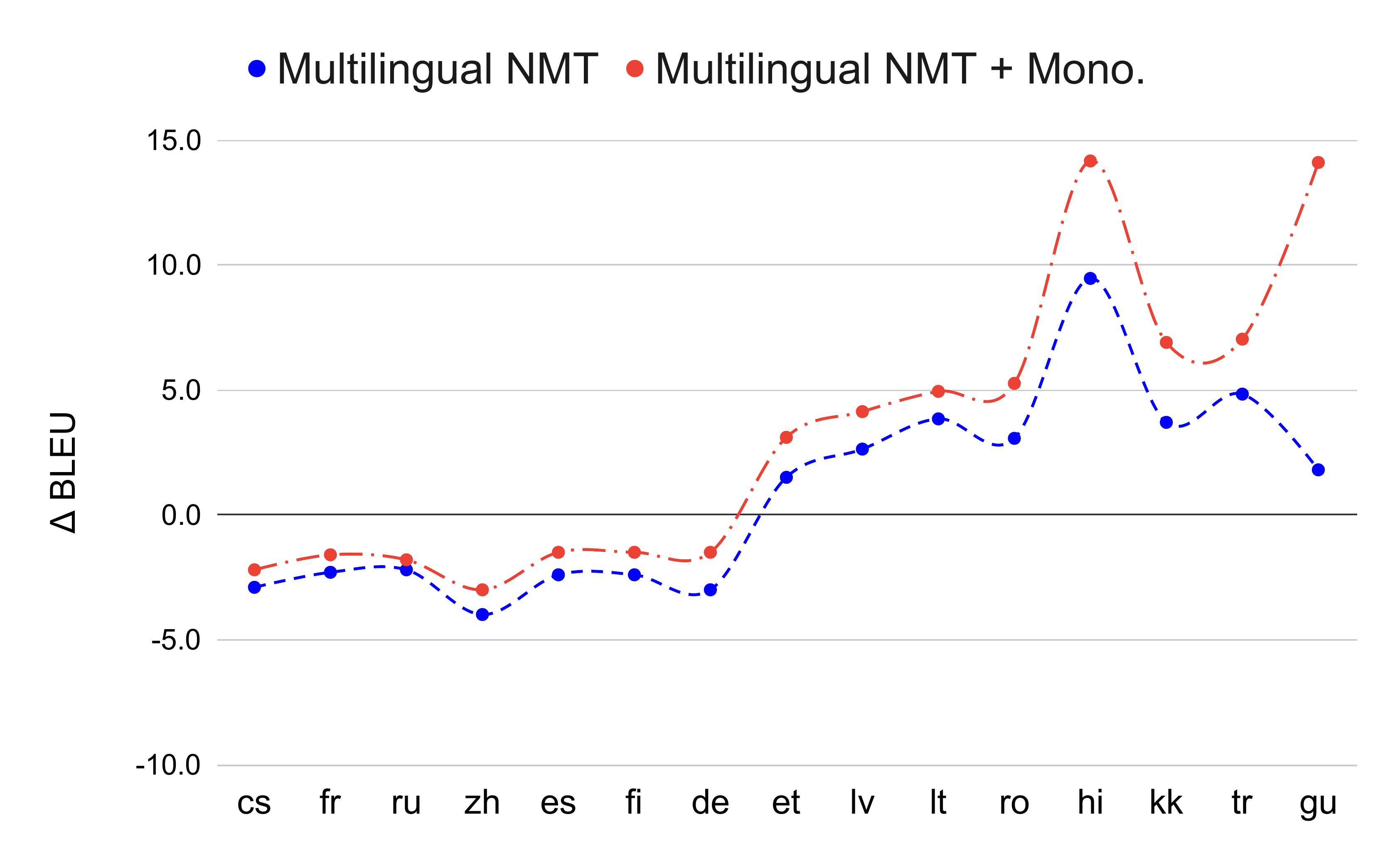} \label{fig:xxen}}
  \subfloat[English-to-Any (en $\rightarrow$ xx)]{\includegraphics[width=0.49\textwidth]{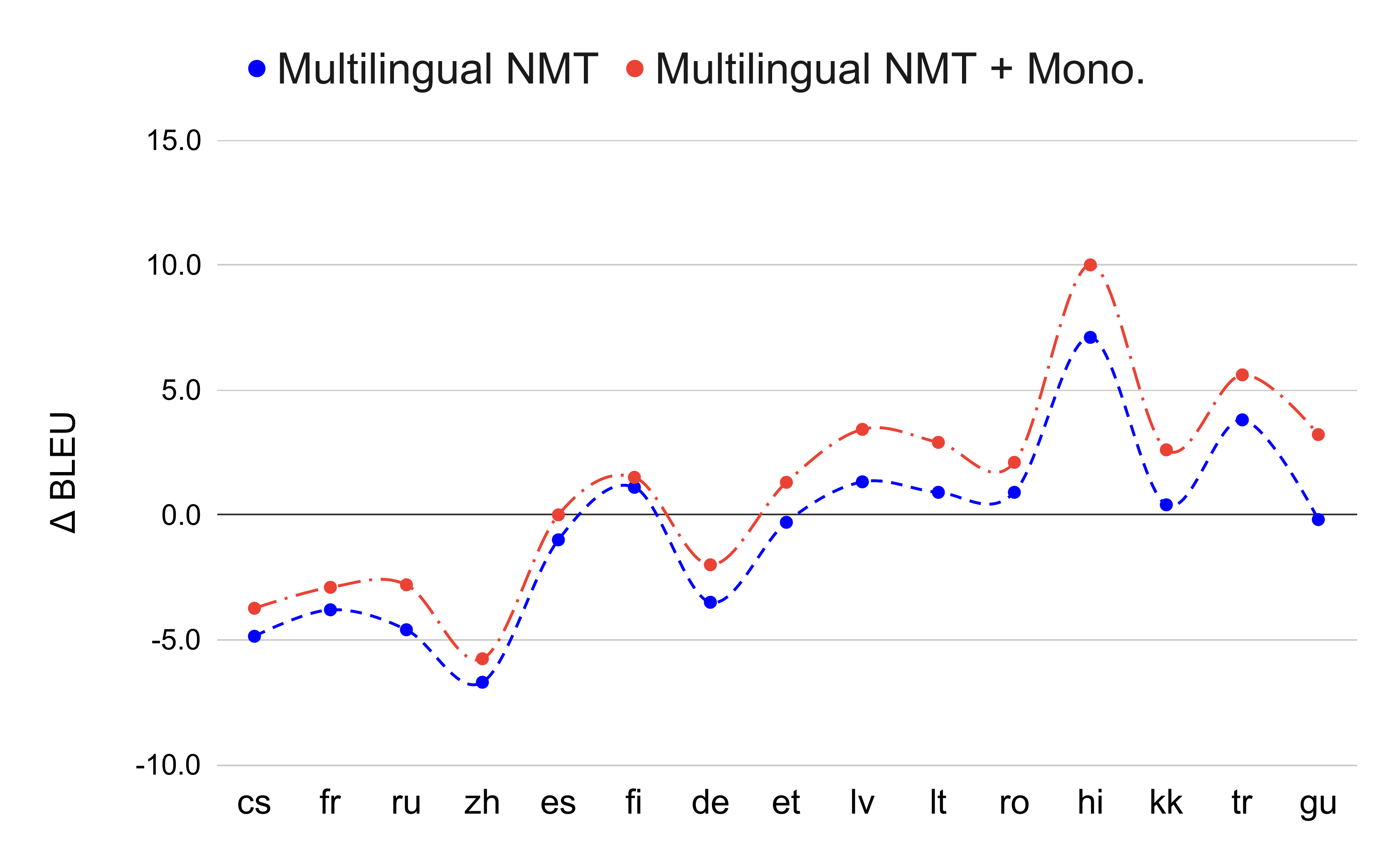} \label{fig:enxx}}
  \caption{Translation quality of Multilingual NMT models relative to bilingual baselines with and without monolingual data. The left plot shows xx $\rightarrow$ en direction and right one shows en $\rightarrow$ xx direction. From left to right on x-axis, we go from high-resource to low-resource languages. The x-axis reflects the bilingual baselines.} \label{fig:multilingual}
\end{figure*}

\subsection{Architecture and Optimization}
All experiments are performed with the Transformer architecture \citep{vaswani2017attention} using the open-source Tensorflow-Lingvo implementation \citep{lingvo}. Specifically, we use the Transformer Big model containing 375M parameters (6 layers, 16 heads, 8192 hidden dimension) \citep{chen2018best} and a shared source-target SentencePiece model (SPM)\footnote{https://github.com/google/sentencepiece} \cite{kudo2018sentencepiece}. We use a vocabulary size of 32k for the bilingual models and 64k for the multilingual models. Different SPMs are trained depending on the set of languages supported by the model.

\section{Using Monolingual Data for Multilingual NMT}
We evaluate the performance of the models using SacreBLEU \cite{post2018call} on standard WMT validation and test sets \cite{papineni2002bleu}. The performance of our bilingual baselines for all 30 English-centric language pairs are reported in Table~\ref{tab:bilingual}. We compare the performance of bilingual models, multilingual models trained with just supervised data for 30 language pairs (15 languages to and from English) and multilingual models trained with a combination of supervised and monolingual data in Figure \ref{fig:multilingual}.

\paragraph{High-Resource Translation}
Our results suggest that a single multilingual model is able to match the quality of individual bilingual models with a gap of less than 2 BLEU points for most high-resource languages, with the exception of Chinese (Zh). The slight quality regression is not surprising, given the large number of languages competing for capacity within the same model \cite{arivazhagan2019massively}. We find that adding additional monolingual data improves the multilingual model quality across the board, even for high-resource language pairs.

\paragraph{Low-Resource Translation}
From Figure \ref{fig:multilingual}, we observe that our supervised multilingual NMT model significantly improves the translation quality for most low and medium-resource languages compared with the bilingual baselines. Adding additional monolingual data leads to an additional improvement of 1-2 BLEU for most medium-resource languages. For the lowest-resource languages like Kazakh (kk), Turkish (tr) and Gujarati (gu), we can see that multilingual NMT alone is not sufficient to reach high translation quality. The addition of monolingual data has a large positive impact on very low resource languages, significantly improving quality over the supervised multilingual model. These improvements range from 3-5 BLEU in the en$\rightarrow$xx direction to more than 5 BLEU for the xx$\rightarrow$en translation.

\begin{table*}[t]
\centering
\begin{tabular}{cccccc}
\toprule
\multicolumn{1}{l}{}                                  & \multicolumn{1}{c|}{}                      & fr\_de & \multicolumn{1}{c|}{de\_fr} & cs\_de & de\_cs \\ \midrule
\multicolumn{1}{c|}{\multirow{5}{*}{4 lang.}}    & \multicolumn{1}{c|}{w/ Parallel Data}      & 27.7   & \multicolumn{1}{c|}{35.3}   & ---    & ---    \\
\multicolumn{1}{c|}{}                                 & \multicolumn{1}{c|}{Translation via Pivot} & 21.9   & \multicolumn{1}{c|}{29.2}   & 20.4   & 19.0   \\ \cline{2-6} 
\multicolumn{1}{c|}{}                                 & \multicolumn{1}{c|}{\citet{arivazhagan2019missing}}    & 20.3   & \multicolumn{1}{c|}{26.0}   & ---    & ---    \\
\multicolumn{1}{c|}{}                                 & \multicolumn{1}{c|}{\citet{kim2019pivot}}       & 17.3   & \multicolumn{1}{c|}{---}    & ---    & 14.1   \\ \cline{2-6} 
\multicolumn{1}{c|}{}                                 & \multicolumn{1}{c|}{Multilingual NMT}          & 11.8   & \multicolumn{1}{c|}{15.2}   & 12.3   & 8.2    \\
\multicolumn{1}{l|}{}                                 & \multicolumn{1}{c|}{Multilingual NMT + Mono.}   & 18.5   & \multicolumn{1}{c|}{27.2}   & 16.9   & 12.6   \\ \midrule
\multicolumn{1}{c|}{\multirow{2}{*}{30 lang.}} & \multicolumn{1}{c|}{Multilingual NMT}          & 10.3   & 14.2                        & 10.5   & 4.3    \\ \cline{2-6} 
\multicolumn{1}{c|}{}                                 & \multicolumn{1}{c|}{Multilingual NMT + Mono.}   & 16.6   & 22.3                        & 14.8   & 7.9    \\ \bottomrule
\end{tabular}
\caption{Zero-shot performance on non-English centric language pairs. We compare with pivot-based translation and two recent approaches from \citet{arivazhagan2019missing} and \citet{kim2019pivot}. The translation quality between these language pairs when parallel data is available is also provided as a baseline. 4 lang. is a multilingual model trained on 4 language pairs (2 languages to and from English), while 30 lang. is our multilingual model trained on all English-centric language pairs.}  \label{tab:zeroshot}
\end{table*}

\begin{table*}[h]
\centering
\resizebox{\linewidth}{!}{
\begin{tabular}{c|cc|cc|cc|cc|cc|cc}
\toprule
                         & fr\_en & en\_fr & de\_en & en\_de & ro\_en & en\_ro & lt\_en & en\_lt & lv\_en & en\_lv & hi\_en & en\_hi \\ \midrule
Multilingual NMT             & 34.9   & 37.5   & 28.7   & 26.4   & 33.2   & 24.3   & 25.1   & 12.4   & 17.6   & 15.5   & 18.0   & 11.6   \\ \midrule
Mono. Only           & 9.8    & 7.6    & 7.4    & 5.8    & 6.8    & 7.3    & 4.8    & 2.1    & 2.9    & 1.8    & 5.3    & 3.1     \\
Multilingual NMT - xx        & 8.4    & 2.4    & 3.9    & 2.6    & 6.2    & 3.8    & 2.2    & 1.1    & 2.1    & 1.7    & 0.8    & 0.6    \\ \midrule
\begin{tabular}[c]{@{}c@{}}Multilingual NMT - xx \\ + Mono.\end{tabular} & 30.7   & 9.8    & 24.2   & 8.9    & 33.0   & 9.3    & 21.3   & 6.7    & 18.8   & 6.1    & 14.6   & 5.4   \\ \bottomrule
\end{tabular}}
\caption{Translation quality of the new language added to Multilingual NMT using just monolingual data. Multilingual NMT here is a multilingual model with 30 language pairs, Mono. Only is a bilingual model used as a baseline trained with only monolingual data with self-supervised learning, Multilingual NMT-xx is a multilingual model trained on 28 language pairs (xx is the language not present in the model). Multilingual NMT-xx + Mono. is a multilingual model with 28 language pairs but only monolingual data for xx.}  \label{tab:unsup}
\end{table*}

\paragraph{Zero-Shot Translation}
We next evaluate the effect of training on additional monolingual data on zero-shot translation in multilingual models. Table~\ref{tab:zeroshot} demonstrates the zero-shot performance of our multilingual model that is trained on 30 language pairs, and evaluated on French(fr)-German(de) and German(de)-Czech(cs), when trained with and without monolingual data. To compare with the existing work on zero-shot translation, we also evaluate the performance of multilingual models trained on just the relevant languages (en-fr-de for fr-de translation, en-cs-de for cs-de translation).
We observe that the additional monolingual data significantly improves the quality of zero-shot translation, often resulting in 3-6 BLEU increase on all zero-shot directions compared to our multilingual baseline. We hypothesize that the additional monolingual data seen during the self-supervised training process helps better align representations across languages, akin to the smoothing effect in semi-supervised learning \cite{Chapelle:2010:SL:1841234}. We leave further exploration of this intriguing phenomenon to future work.

\section{Adding New Languages to Multilingual NMT}
\vspace{-10px}
Inspired by the effectiveness of monolingual data in boosting low-resource language translation quality, we continue with a stress-test in which we completely remove the available parallel data from our multilingual model, one language at a time, in order to observe the unsupervised machine translation quality for the missing language.

Results of this set of experiments are detailed in Table~\ref{tab:unsup}. We find that simply adding monolingual data for a new language to the training procedure of a multilingual model is sufficient to obtain strong translation quality for several languages, often attaining within a few BLEU points of the fully supervised multilingual baseline, without the need for iterative back-translation. We also notice significant quality improvements over models trained with just self-supervised learning using monolingual data for a variety of languages. On WMT ro-en, the performance of our model exceeds XLM \cite{conneau2019cross} by over 1.5 BLEU and matches bilingual MASS \cite{song2019mass}, without utilizing any back-translation. This suggests that jump-starting the iterative back-translation process from multilingual models might be a promising avenue to supporting new languages.

\section{Related Work}
\vspace{-10px}
Our work builds on several recently proposed techniques for multilingual NMT and self-supervised representation learning. While massively multilingual models have obtained impressive quality improvements for low-resource languages as well as zero-shot scenarios \cite{aharoni2019massively,arivazhagan2019missing}, it has not yet been shown how these massively multilingual models could be extended to unseen languages, beyond the pipelined approaches \cite{currey2019zero,lakew2019adapting}. On the other hand, self-supervised learning approaches have excelled at down-stream cross-lingual transfer \cite{devlin2019bert,raffel2019exploring,conneau2019unsupervised}, but their success for unsupervised NMT \cite{conneau2019cross,song2019mass} currently lacks robustness when languages are distant or monolingual data domains are mismatched \cite{neubig2018rapid,vulic2019we}. We observe that these two lines of research can be quite complementary and can compensate for each other's deficiencies.

\section{Conclusion and Future Directions}
\vspace{-10px}
We present a simple framework to combine multilingual NMT with self-supervised learning, in an effort to jointly exploit the learning signals from multilingual parallel data and monolingual data. We demonstrate that combining multilingual NMT with monolingual data and self-supervision (i) improves the translation quality for both low and high-resource languages in a multilingual setting, (ii) leads to on-par zero-shot capability compared with competitive bridging-based approaches and (iii) is an effective way to extend multilingual models to new unseen languages.

Future work should explore techniques like iterative back-translation \cite{hoang2018iterative} for further improvement and scaling to larger model capacities and more languages \cite{arivazhagan2019massively,huang2019gpipe} to maximize transfer across languages and across data sources.

\bibliography{anthology,acl2020}
\bibliographystyle{acl_natbib}

\appendix

\begin{table*}[h]
\centering
\label{table:dataset_bi}
\begin{tabular}{>{\centering\arraybackslash}m{2cm}>{\centering\arraybackslash}m{1.7cm}>{\centering\arraybackslash}m{1.7cm}>{\centering\arraybackslash}m{1.7cm}>{\centering\arraybackslash}m{1.7cm}>{\centering\arraybackslash}m{1.2cm}>{\centering\arraybackslash}m{1.2cm}}
\toprule
 \multirow{2}{1.5cm}{\centering Language Pair} & \multicolumn{3}{c}{Data Sources} & \multicolumn{3}{c}{$\#$ Samples}\\
 \cmidrule{2-7}
  & Train & Dev & Test & Train & Dev & Test\\
\midrule
cs$\rightarrow$en                                                            & WMT'19    & WMT'17    & WMT'18   & 64336053     & 3005    & 2983    \\
fr$\rightarrow$en                                                            & WMT'15    & WMT'13    & WMT'14   & 40449146     & 3000    & 3003    \\
ru$\rightarrow$en                                                            & WMT'19    & WMT'18    & WMT'19   & 38492126     & 3000    & 2000    \\
zh$\rightarrow$en                                                            & WMT'19    & WMT'18    & WMT'19   & 25986436     & 3981    & 2000    \\
es$\rightarrow$en                                                            & WMT'13    & WMT'13    & WMT'13   & 15182374     & 3004    & 3000    \\
fi$\rightarrow$en                                                            & WMT'19    & WMT'18    & WMT'19   & 6587448      & 3000    & 1996    \\
de$\rightarrow$en                                                            & WMT'14    & WMT'13    & WMT'14   & 4508785      & 3000    & 3003    \\
et$\rightarrow$en                                                            & WMT'18    & WMT'18    & WMT'18   & 2175873      & 2000    & 2000    \\
lv$\rightarrow$en                                                            & WMT'17    & WMT'17    & WMT'17   & 637599       & 2003    & 2001    \\
lt$\rightarrow$en                                                            & WMT'19    & WMT'19    & WMT'19   & 635146       & 2000    & 1000    \\
ro$\rightarrow$en                                                            & WMT'16    & WMT'16    & WMT'16   & 610320       & 1999    & 1999    \\
hi$\rightarrow$en                                                            & WMT'14    & WMT'14    & WMT'14   & 313748       & 520     & 2507    \\
kk$\rightarrow$en                                                            & WMT'19    & WMT'19    & WMT'19   & 222424       & 2066    & 1000    \\
tr$\rightarrow$en                                                            & WMT'18    & WMT'17    & WMT'18   & 205756       & 3007    & 3000    \\
gu$\rightarrow$en                                                            & WMT'19    & WMT'19    & WMT'19   & 11670        & 1998    & 1016    \\
\midrule
en$\rightarrow$cs                                                            & WMT'19    & WMT'17    & WMT'18   & 64336053     & 3005    & 2983    \\
en$\rightarrow$fr                                                            & WMT'15    & WMT'13    & WMT'14   & 40449146     & 3000    & 3003    \\
en$\rightarrow$ru                                                            & WMT'19    & WMT'18    & WMT'19   & 38492126     & 3000    & 2000    \\
en$\rightarrow$zh                                                            & WMT'19    & WMT'18    & WMT'19   & 25986436     & 3981    & 2000    \\
en$\rightarrow$es                                                            & WMT'13    & WMT'13    & WMT'13   & 15182374     & 3004    & 3000    \\
en$\rightarrow$fi                                                            & WMT'19    & WMT'18    & WMT'19   & 6587448      & 3000    & 1996    \\
en$\rightarrow$de                                                            & WMT'14    & WMT'13    & WMT'14   & 4508785      & 3000    & 3003    \\
en$\rightarrow$et                                                            & WMT'18    & WMT'18    & WMT'18   & 2175873      & 2000    & 2000    \\
en$\rightarrow$lv                                                            & WMT'17    & WMT'17    & WMT'17   & 637599       & 2003    & 2001    \\
en$\rightarrow$lt                                                            & WMT'19    & WMT'19    & WMT'19   & 635146       & 2000    & 1000    \\
en$\rightarrow$ro                                                            & WMT'16    & WMT'16    & WMT'16   & 610320       & 1999    & 1999    \\
en$\rightarrow$hi                                                            & WMT'14    & WMT'14    & WMT'14   & 313748       & 520     & 2507    \\
en$\rightarrow$kk                                                            & WMT'19    & WMT'19    & WMT'19   & 222424       & 2066    & 1000    \\
en$\rightarrow$tr                                                            & WMT'18    & WMT'17    & WMT'18   & 205756       & 3007    & 3000    \\
en$\rightarrow$gu                                                            & WMT'19    & WMT'19    & WMT'19   & 11670        & 1998    & 1016    \\
\midrule
fr$\rightarrow$de                                                            & WMT'19    & WMT'13    & WMT'13   & 9824476      & 1512    & 1701    \\
de$\rightarrow$fr                                                            & WMT'19    & WMT'13    & WMT'13   & 9824476      & 1512    & 1701    \\
cs$\rightarrow$de                                                            & ----      & WMT'13    & WMT'13   & ---          & 1997    & 1997    \\
de$\rightarrow$cs                                                            & ----      & WMT'13    & WMT'13   & ---          & 1997    & 1997    \\
\bottomrule
\end{tabular}
\caption{Data sources and number of samples for the parallel data in our corpus. Please note that we don't use parallel data in Fr-De for any of the experiments in the paper apart from training parallel data baseline in Table 2. We don't have any parallel data in Cs-De. } 
\end{table*}

\appendix

\begin{table*}[h]
\centering
\label{table:datasets_mono}
\begin{tabular}{>{\centering\arraybackslash}m{1.4cm}>{\centering\arraybackslash}m{1cm}>{\centering\arraybackslash}m{1.5cm}>{\centering\arraybackslash}m{1.3cm}>{\centering\arraybackslash}m{1.1cm}>{\centering\arraybackslash}m{1.4cm}>{\centering\arraybackslash}m{1.2cm}>{\centering\arraybackslash}m{1.7cm}>{\centering\arraybackslash}m{0.9cm}>{\centering\arraybackslash}m{0.9cm}}
\toprule
\multirow{2}{1.4cm}{\centering Language} & \multicolumn{6}{c}{\centering Data Sources} & \multicolumn{3}{c}{$\#$ Samples}\\
\cmidrule{2-10}
& News Crawl & News Commentary & Common Crawl & Europarl & News Discussions & Wiki Dumps & Train & Dev & Test\\
\midrule
en & \checkmark & & & & & & 199900557 & 3000 & 3000\\
ro & \checkmark & & & & & & 14067879 & 3000 & 3000\\
de & \checkmark & & & & & & 275690481 & 3000 & 3000\\
fr & \checkmark & \checkmark & & \checkmark & \checkmark & & 160933435 & 3000 & 3000\\
cs & \checkmark & & & & & & 72157988 & 3000 & 3000\\
es & \checkmark & & & & & & 43814290 & 3000 & 3000\\
et & \checkmark & & \checkmark & & & & 51683012 & 3000 & 3000\\
fi & \checkmark & & & \checkmark & & & 18847600 & 3000 & 3000\\
gu & \checkmark & & \checkmark & & & & 4644638 & 3000 & 3000\\
hi & \checkmark & & & & & & 23611899 & 3000 & 3000\\
kk & \checkmark & \checkmark & \checkmark & & & \checkmark & 13825470 & 3000 & 3000\\
lt & \checkmark & & \checkmark & \checkmark & & \checkmark & 106198239 & 3000 & 3000\\
lv & \checkmark & & & \checkmark & & & 10205015 & 3000 & 3000\\
ru & \checkmark & & & & & & 80148714 & 3000 & 3000\\
tr & \checkmark & & & & & & 9655009 & 3000 & 3000\\
zh & \checkmark & \checkmark & & & & & 2158309 & 3000 & 3000\\
\bottomrule
\end{tabular}
\caption{Data sources and number of samples for the monolingual data in our corpus.} 
\end{table*}

\appendix

\section{Appendices}

\begin{table*}[]
\centering
\begin{tabular}{c|cccc}
\toprule
Language Pair     & \begin{tabular}[c]{@{}c@{}}Bilingual\\ Baseline\end{tabular} & \begin{tabular}[c]{@{}c@{}}Multilingual\\ NMT\end{tabular} & \begin{tabular}[c]{@{}c@{}}Multilingual\\ NMT + Mono.\end{tabular} & SOTA \\
\midrule
cs$\rightarrow$en & 29.7                                                                                                                                  & 28.4         & 29.1                                                                                                                                 & 33.9 \\
fr$\rightarrow$en & 35.5                                                                                                                                  & 34.9         & 35.6                                                                                                                                 & 39.5 \\
ru$\rightarrow$en & 34.9                                                                                                                                  & 33.8         & 34.1                                                                                                                                 & 40.1 \\
zh$\rightarrow$en & 21.7                                                                                                                                  & 17.7         & 18.7                                                                                                                                 & 39.3 \\
es$\rightarrow$en & 30.1                                                                                                                                  & 28.9         & 29.6                                                                                                                                 & 31.4 \\
fi$\rightarrow$en & 26.0                                                                                                                                  & 25.2         & 25.8                                                                                                                                 & 33.0 \\
de$\rightarrow$en & 27.4                                                                                                                                  & 27.2         & 28.1                                                                                                                                 & 32.0 \\
et$\rightarrow$en & 24.3                                                                                                                                  & 24.2         & 24.9                                                                                                                                 & 30.9 \\
lv$\rightarrow$en & 15.0                                                                                                                                  & 17.6         & 18.8                                                                                                                                 & 36.3 \\
lt$\rightarrow$en & 21.3                                                                                                                                  & 24.4         & 25.4                                                                                                                                 & 36.3 \\
ro$\rightarrow$en & 30.1                                                                                                                                  & 33.0         & 34.1                                                                                                                                 & 38.5 \\
hi$\rightarrow$en & 8.5                                                                                                                                   & 16.0         & 18.5                                                                                                                                 & 16.7 \\
kk$\rightarrow$en & 4.7                                                                                                                                   & 11.2         & 17.6                                                                                                                                 & 30.5 \\
tr$\rightarrow$en & 15.9                                                                                                                                  & 18.4         & 21.1                                                                                                                                 & 28.0 \\
gu$\rightarrow$en & 2.0                                                                                                                                   & 3.0          & 15.1                                                                                                                                 & 24.9 \\
\midrule
en$\rightarrow$cs & 23.8                                                                                                                                  & 20.0         & 20.3                                                                                                                                 & 29.9 \\
en$\rightarrow$fr & 38.1                                                                                                                                  & 36.2         & 36.6                                                                                                                                 & 43.8 \\
en$\rightarrow$ru & 24.9                                                                                                                                  & 22.0         & 22.9                                                                                                                                 & 36.3 \\
en$\rightarrow$zh & 31.3                                                                                                                                  & 5.0          & 5.9                                                                                                                                  & 36.3 \\
en$\rightarrow$es & 32.8                                                                                                                                  & 29.7         & 30.0                                                                                                                                 & 30.4 \\
en$\rightarrow$fi & 20.3                                                                                                                                  & 19.2         & 19.6                                                                                                                                 & 27.4 \\
en$\rightarrow$de & 26.4                                                                                                                                  & 22.1         & 23.9                                                                                                                                 & 27.1 \\
en$\rightarrow$et & 19.0                                                                                                                                  & 18.9         & 20.1                                                                                                                                 & 25.2 \\
en$\rightarrow$lv & 14.2                                                                                                                                  & 14.9         & 16.5                                                                                                                                 & 21.1 \\
en$\rightarrow$lt & 11.0                                                                                                                                  & 10.9         & 14.4                                                                                                                                 & 20.1 \\
en$\rightarrow$ro & 23.7                                                                                                                                  & 23.6         & 24.8                                                                                                                                 & 33.3 \\
en$\rightarrow$hi & 4.5                                                                                                                                   & 10.6         & 13.9                                                                                                                                 & 12.5 \\
en$\rightarrow$kk & 0.2                                                                                                                                   & 1.1          & 4.3                                                                                                                                  & 11.1 \\
en$\rightarrow$tr & 13.7                                                                                                                                  & 13.8         & 15.7                                                                                                                                 & 20.0 \\
en$\rightarrow$gu & 0.6                                                                                                                                   & 0.4          & 4.0                                                                                                                                  & 28.2 \\ \bottomrule
\end{tabular}
\caption{Absolute BLEU scores for results in Figure 2 in the paper.}
\end{table*}

\end{document}